\documentclass[letterpaper, 10 pt, conference]{ieeeconf}  

\IEEEoverridecommandlockouts                              
\usepackage{graphics} 
\usepackage{epsfig} 
\usepackage{times} 
\usepackage{amsmath} 
\usepackage{amssymb}  
\usepackage{cite}
\usepackage{amsfonts}
\usepackage{amsthm}
\usepackage{enumerate}
\newtheorem{lem}{Lemma}
\usepackage{xcolor}
\usepackage{kky}
\usepackage{cleveref}
\usepackage{subfigure}
\usepackage[ruled,linesnumbered]{algorithm2e}
\newtheorem{remk}{Remark}
\newcommand{\ar}[1]{{\color{red} Arash: #1}}
\newcommand{\an}[1]{{\color{blue} Andrea: #1}}
\newcommand{\ours}{ODE\&ROA\xspace}

\renewcommand{\ar}[1]{{\color{red}  {}}}
\renewcommand{\an}[1]{{\color{blue}  {}}}

\title{\LARGE \bf
Learning Dynamical Systems
using Local Stability Priors
}

\author{Arash Mehrjou$^{1}$, Andrea Iannelli$^{2}$, and Bernhard Sch\"olkopf$^{3}$
\thanks{}
\thanks{$^{1}$A. Mehrjou is with the Empirical Inference Department of Max Planck Institute for Intelligent Systems, T\"ubingen, Germany, and Department of Computer Science at ETH Z\"urich. Faculty of Electrical Engineering, Z\"urich, Switzerland.
        {\tt\small amehrjou@tue.mpg.de}}%
\thanks{$^{2}$Andrea Iannelli is with the Department of Electrical Engineering, Automatic Control Lab, ETH, Z\"{u}rich 8092, Switzerland
{\tt\small iannelli@control.ee.ethz.ch}}
\thanks{$^{3}$B. Sch\"olkopf is with the Empirical Inference Department of Max Planck Institute for Intelligent Systems, T\"ubingen, Germany.
        {\tt\small bs@tue.mpg.de}}%
}

\begin{document}

\maketitle
\thispagestyle{empty}
\pagestyle{empty}

\begin{abstract}

   A coupled computational approach to simultaneously learn a vector field and the region of attraction of an equilibrium point from generated trajectories of the system is proposed. The nonlinear identification leverages the local stability information as a prior on the system, effectively endowing the estimate with this important structural property. In addition, the knowledge of the region of attraction plays an experiment design role by informing the selection of initial conditions from which trajectories are generated and by enabling the use of a Lyapunov function of the system as a regularization term. Numerical results show that the proposed method allows efficient sampling and provides an accurate estimate of the dynamics in an inner approximation of its region of attraction.

\end{abstract}

\section{Introduction}
Learning ordinary differential equations (ODE) of a dynamical system given observed trajectories is the main goal of system identification~\cite{Ljung1999}. To achieve a satisfactory accuracy, it is recognized the importance, especially in a nonlinear context, of exploiting prior knowledge of the system ~\cite{Ljung_NL}. A distinctive property due to nonlinearity is that stability is no more a feature associated with the whole system, as in the linear realm, but with each of its attractors~\cite{Khalil_book}. In fact, stability becomes now in the general case a local attribute, which only holds in regions surrounding the attractor. This is the case, for example, for the region of attraction (ROA) of equilibria~\cite{Chesi_book} and region of contraction of limit cycles~\cite{Giesl}.

Given its specificity and importance in describing the qualitative and quantitative properties of a system, local stability priors seem an important structural constraint to encode in a nonlinear identification algorithm. While this has been already done for identification of linear systems (with e.g. subspace methods~\cite{StableSubSpace}, maximum likelihood~\cite{Umenberger_stab}), it is a new idea, to the best of the author's knowledge, in learning nonlinear ODEs.
On the other hand, computing this type of priors is notoriously a difficult task~\cite{Genesio_ROA} and there are not state-of-practice analytical methods capable of handling generic problems.

Starting from these premises, the work develops an entirely data-driven learning and stability analysis framework that, given access to trajectories of the true system for user-specified initial conditions, focuses on one attractor and iteratively builds up estimates of both its local stability region and the dynamics (i.e. an approximate ODE describing it) in that region. These seemingly distinct learning processes are jointly executed and coupled via encoding priors.

As a first step towards a general framework that is able to handle rich dynamics comprising different types of attractors, here we will only consider dynamical systems with one or more locally attractive equilibrium points. Once a certain equilibrium has been selected, the estimated ODE and ROA are only associated with it (that is, a new algorithm should be run and new estimates would be obtained for a different fixed point). While this is an obvious result for ROA since, as stated before, stability features are associated with the single attractor, the fact that multiple estimates of the vector field (one for each equilibrium) are obtained might be at first unconvincing. This can, however, find several motivations. From an application perspective, it is reasonable to seek the best approximation in the region where the system will be deployed (e.g. nonlinear aircraft~\cite{Iannelli_TCST18} or power systems~\cite{NL_powerSys} models). From a system identification perspective, this strategy is believed to enable the originally difficult task of identifying a full nonlinear system by structuring it as the identification of a model for each attractor. Further implications and benefits of this \emph{ad-attractor} learning strategy will be detailed in the relevant technical sections.

\emph{Related work:}
While most of the approaches in the ROA literature are model-based \cite{Chesi_book} and research is still active in that field \cite{ValmorbidaAuto2017,Iannelli_Auto_IQC,ROA_recent}, recent works have also considered purely data-driven methods.
A sampling strategy is proposed in~\cite{najafi2016fast} to estimate the ROA of a system in real-time. A probabilistic method is used in~\cite{berkenkamp2016safe} to safely sample and learn the ROA for systems with uncertainty. In a preliminary effort towards ROA estimation, the authors proposed in~\cite{mehrjou2019deep} a deep network architecture combined with a sampling strategy to discover a Lyapunov function for a known system. In~\cite{richards2018lyapunov} a Lyapunov function was learned by a neural network and gradually changed such that its level sets become closer to the ROA of the system.
A simulation-based approach was also investigated in~\cite{Topcu_Auto_Sim}, where however the computation relied on Sum of Square optimization, typically the cornerstone of model-based ROA methods.

Using machine learning to learn ODEs has also been explored using various learning tools. Kernel methods \cite{scholkopf2002learning} have been extensively used for time-series prediction and ODE estimation~\cite{michalak2011time}. In a slightly different context, Fuzzy and Neuro-Fuzzy methods have been successfully applied to nonlinear systems with uncertainty~\cite{jafari2016fuzzy}.
The deep neural network is another nonlinear function approximating technique that has been adopted for ODE learning, but its opaqueness has made it less interesting for the community of system identification and control. However, a multi-layer perceptron as a universal function approximator can still be used to capture the nonlinear dynamics~\cite{rubanova2019latent}. Recently, a new class of neural networks has been proposed where the time steps are modeled as the layers of the network. Hence, continuous dynamics correspond to a network with infinitely many layers~\cite{chen2018neural}.

Recently,\cite{mehrjou2020automatic} proposed an iterative technique to learn an improving sequence of controllers that enlarge the ROA of the system. Here, we propose a similar iterative method while the goal is system identification instead of control synthesis.

\emph{Contribution:}
An iterative learning strategy whereby ODE and ROA of a dynamical system are learned from observed trajectories is proposed. Multilayer perceptrons are used as universal function approximators for both the Lyapunov function approximating the true ROA and the estimate of the vector field of the system. Learning ROA and ODE are interlaced and inform each other through an iterative algorithm that is shown to be more efficient than learning them separately. The main advantage of co-learning ROA and ODE is twofold. First, the former can be used to frugally sample initial conditions for the (numerical or real) experiments. Second, learning the second can be regularized such that the ODE learning process is biased towards dynamical systems for which level sets of the current Lyapunov function are inner estimates of the true ROA. The proposed method is supported by experiments on the known Van der Pol oscillator benchmark.

\section{Co-learning ROA and ODE}

\subsection{Problem statement}
\label{sec:problem_statement}
Consider an autonomous nonlinear system of the form:
\begin{equation}\label{eq:auton_sys}
\dot{x}=f(x), \quad x(0)=x_0,
\end{equation}
where $f: \mathbb{R}^n \rightarrow \mathbb{R}^n$ is called vector field.
The vector $\bar{x} \in \mathbb{R}^n$ is an equilibrium point of~\eqref{eq:auton_sys} if $f(\bar{x})=0$.
Let $\phi(t,x_0)$ denote the associated flow, i.e. the solution of~\eqref{eq:auton_sys} at time $t$ with initial condition $x_0$. The region of attraction (ROA) associated with $\bar{x}$ is defined as:
\begin{equation}\label{ROA_def}
\mathcal{R}_{\bar{x}}:=\big\{x_0 \in \mathbb{R}^n: \lim_{t\to\infty} \phi(t,x_0)= \bar{x}  \big\}.
\end{equation}
That is, $\mathcal{R}_{\bar{x}}$ is the set of all initial states that eventually converge to $\bar{x}$.
While for linear systems convergence to the equilibrium is a global property independent of the equilibrium (which is always one), for nonlinear ones it might hold only locally, thus $\mathcal{R}_{\bar{x}} \subseteq \mathbb{R}^n$. The goal of the work is to propose a data-driven approach to jointly estimate the vector field $f_{\bar{x}}$ from trajectories originating from the basin of attraction of $\bar{x}$ and its ROA $\mathcal{R}_{\bar{x}}$. The dependence of both on the considered equilibrium is emphasized with the subscript (which later on will be dropped when unessential for the sake of a lighter notation).

In practice, trajectories will be measured at discrete times, which are assumed to be contaminated by a zero-mean i.i.d. Gaussian noise with variance $\sigma^2$. Starting from (\ref{eq:auton_sys}), the problem can then be equivalently represented as:
\begin{subequations}\label{eq:dynamical_system_discrete}
\begin{align}
x_{t+1}&=f(x_t), \label{eq:dynamical_system_discrete_1}\\
y_t&=x_t+\epsilon, \quad \epsilon \sim \Ncal(0, \sigma^2),\label{eq:dynamical_system_discrete_2}
\end{align}
\end{subequations}
where $t\in \NN$ is a time index.

Let denote by $x_{0:T}$ the set of states obtained by sampling the flow $\phi(t,x_0)$ in the set of time indexes $\{0, 1, \ldots, T\}$, that is $x_{0:T}=\{x_0, x_1, \ldots, x_T\}$. 
Using as inputs these data sets (contaminated by noise as described in Eq. \ref{eq:dynamical_system_discrete_2}), two machine learning algorithms, namely neural networks and kernel methods, will be used to compute an estimate $\hat{f}_{\bar{x}}$.
It is at this point important to observe that each set $x_{0:T}$ is associated with a different initial condition $x_0$, and in general, the total number of available trajectories is limited by a defined budget, which has to do with e.g. cost of experiments, the time required for simulation. Besides the quantity, also the locations of the initial conditions might be constrained by, e.g. physical limitations (the system has to be safely operated), the validity of the model (which is only ensured in certain regions of the state space). It is thus important to recognize that $x_0$ plays in this input-free setting the role of \emph{external} action to probe the dynamics. In other words, each $x_0$ will generate a different trajectory which in turn, due to nonlinearity, will have different informativity. Effectively, the selection of $x_0$ is an \emph{experiment design} problem in this setting.
Leveraging the impact of distinct initial conditions in nonlinear identification problems is indeed a key enabler in this work, and it has not been fully exploited in the literature (see for a notable exception the recent work in \cite{Korda_opt_K} where the rich spectrum of nonlinear responses associated with different initial conditions is leveraged in the context of identifying the Koopman operator).

Frugality on one hand (i.e. estimating $f_{\bar{x}}$ using the least possible sets $x_{0:T}$) and safety and informativity on the other, make the region $\mathcal{R}_{\bar{x}}$
of the state space an ideal candidate to draw initial conditions from.

Standard approaches to compute inner approximations $\mathcal{\hat{R}}_{\bar{x}}$ of the true ROA $\mathcal{R}_{\bar{x}}$ are model-based, i.e. they require knowledge of $f$. While this is not the case here, results from the literature can still be leveraged. Specifically, it will be used the known fact that Lyapunov function level sets provide inner estimates of $\mathcal{R}_{\bar{x}}$~\cite{Khalil_book}.
\begin{lem}\label{lemma:LF_LevelSets}
Let $\mathcal{D} \subset \mathbb{R}^n$ and let $\bar{x} \in \mathcal{D}$. If there exists a $1$-time continuously differentiable function $V: \mathbb{R}^n \rightarrow \mathbb{R}$ such that:
\begin{equation}\label{eq:Lyap_theorem}
\begin{aligned}
&V(\bar{x})=0 \quad \textnormal{and} \quad V(x)>0 \quad \quad \forall x \in \mathcal{D}\backslash \bar{x},\\
&  \nabla V(x)f(x)<0 \quad \hspace{0.85in} \forall x \in \mathcal{D}\backslash \bar{x},\\
& \Vcal_{\gamma}(\theta):=\{x \in \mathbb{R}^n: V(x)\leq \gamma  \}, \hspace{0.13in} \Vcal_{\gamma} \subseteq \mathcal{D}, \\
\end{aligned}
\end{equation}
and $\Vcal_{\gamma}$ is bounded. Then, $\mathcal{\hat{R}}_{\bar{x}}$=$\Vcal_{\gamma} \subseteq \mathcal{R}_{\bar{x}}$.
\end{lem}
When $f$ is known, a common approach to compute $\mathcal{\hat{R}}$ is via Sums of Squares (SOS) optimization \cite{Parrilo_MP2003,Chesi_book}, thereby one finds polynomial functions that satisfy set containment conditions (as those in Eq. \ref{eq:Lyap_theorem}).

In this work, $\mathcal{\hat{R}}_{\bar{x}}$ is co-learned with
$f_{\bar{x}}$ by looking for Lyapunov functions with contractive level sets (\ref{eq:Lyap_theorem}). Depending on the chosen degree of the polynomial Lyapunov function, the estimates in SOS-based approaches can be quite conservative, i.e., $\vol(\mathcal{\hat{R}}_{\bar{x}}) \ll \vol(\mathcal{R}_{\bar{x}})$. To overcome this issue, the universal approximation property of deep neural networks is exploited to compute Lyapunov functions whose largest contractive level set well approximates the shape of $\mathcal{R}_{\bar{x}}$  (Section \ref{sec:roa_estimation}).

The estimated ROA is then used to learn $f_{\bar{x}}$ from trajectories originated within the $\mathcal{R}_{\bar{x}}$. Besides this experiment design role, the knowledge of $\mathcal{R}_{\bar{x}}$ is also exploited to regularize the learning algorithm for $f_{\bar{x}}$ by an appropriate local stability priors (Section \ref{sec:ode_estimation}).
These two learning stages are coupled together and, in a way, symbiotic since both benefit from the other (Section \ref{sec:learning_together_ode_roa}). As the estimated ROA increases in size, a better approximation of the vector field will be achieved while always using a limited (but informative) number of trajectories, no matter the state space region from which these are drawn. Once some predefined convergence criteria are met, this iterative procedure is able to provide the sought estimates $f_{\bar{x}}$ and $\mathcal{R}_{\bar{x}}$.

\subsection{Estimating ROA from trajectories}
\label{sec:roa_estimation}
Let $V(\cdot;\theta):\RR^n\to\RR_+$ be a candidate Lyapunov function for~\eqref{eq:dynamical_system_discrete_1} parameterized by $\theta$. To ensure the $V(\cdot;\theta)$ is positive definite, rather parameterizing it directly, it is modelled as the inner product of a feature extractor with itself, i.e., $V(\cdot;\theta) = v\tran(\cdot;\theta)v(\cdot;\theta)$ where $v(\cdot;\theta):\RR^n \to \RR^n$ is a multilayer perceptron.
We will denote by $\Vcal_c(\theta):=\{x\in\RR^n : V(x;\theta)\leq c\}$ level set of $V$ parametrized with $\theta$ with size $c$.

The goal is then to find $\theta$ and $c$ such that $\Vcal_c(\theta)$ is a good approximation of the true ROA $\mathcal{R}_{\bar{x}}$. The parameters $\theta$ defines the shape of the level sets while $c$ determines its size. Recalling the definition of the ROA (\ref{ROA_def}) and the result in Lemma \ref{lemma:LF_LevelSets}, the following multi-step supervised learning approach is considered.
\begin{enumerate}
        \item The initialization step consists of training $V(\cdot;\theta)$ such that it takes a quadratic shape using the following loss function:
        \begin{equation}
                \label{eq:pretraining_quadratic_loss}
                \theta_0^* = \argmin_\theta \EE_\Bcal[V(x;\theta) - x\tran Q x]
        \end{equation}
        where $\EE_\Bcal$ shows the empirical expectation on the samples uniformly taken from the set $\Bcal$ which is a small ball around the equilibrium. The matrix $Q$ characterizes the target quadratic function which is usually set to $I$.
        A small value of $c$ is then chosen such that $\Vcal_c(\theta_0^*)$ is inside the ROA.
        \item The level set $\Vcal_c$ is expanded gradually by multiplying $c$ with $\alpha>1.0$ and a gap region $\Gcal=\Vcal_{\alpha c} \backslash \Vcal_{c}$ is generated.
        \item The system is simulated by drawing $J$ initial conditions inside $\Gcal$ for $T$ steps. For notational convenience, the $j^{\rm th}$ trajectory is denoted by $\tau^j=x^j_{0:T}$. Each initial state is given the label $+1$ (stable) if its trajectory enters $\Vcal_c$. Otherwise, it is given the label $-1$ (unstable). Let $l:\Gcal\to \{+1, -1\}$ be the oracle function that simulates the system from an initial condition inside the gap $\Gcal$ and assigns the label as mentioned. Ultimately, it produces a dataset consisting of $J$ pairs $\{(\tau^j, l(x_0^j))\}_{j=1}^{j=J}$
        \item $\theta'\leftarrow \theta + \gamma \nabla_\theta \Lcal_\theta$ where

        \begin{align}
                \Lcal_\theta = &\sum_{x_0\in\Gcal} \ell(V(x_0;\theta), l(x_0)) + \\ &\lambda_\theta 1_{[l(x_0)=+1]}\nonumber
                \sum_{t=1}^T [V(x_{t+1};\theta)-V(x_{t};\theta)]
        \end{align}
        and $\ell: \RR\times \{0, 1\}\to \RR$ is defined as
        \begin{align}\label{Loss}
                \ell(V(x;\theta), l(x)) = l(x) [V(x;\theta) - c].
        \end{align}
        This loss function is defined such that the optimized shape of the new level sets $\Vcal(\cdot;\theta')$ will incorporate the states from $\Gcal$ which are labeled $+1$ and exclude those which are labeled $-1$.
        \item Update the value of $c$ by line search so that every $x\in \Vcal_c(\theta')$ satisfies $V(f(x)) - V(x)<0$ as the second condition of Lemma \ref{lemma:LF_LevelSets}. Since in practice $f$ is not known, an empirical version of this condition is tested for the states along the observed trajectories as $V(x_{t+1}) - V(x_t)<0$. Hence, $c$ is set to a value such that the decrease condition is satisfied for all pairs $(x_t, x_{t+1})$ chosen from the observed trajectories and $x_t, x_{t+1}\in\Vcal_c(\theta')$
        \item Go back to step $2$ until no point from $G$ is labelled $+1$.
\end{enumerate}

\begin{remk}
 The proposed loss function (\ref{Loss}) is a key part of the algorithm for learning ROA and is defined by leveraging the result from Lemma \ref{lemma:LF_LevelSets} on contractiveness of Lyapunov function level sets. In the same spirit, other (possibly less conservative) conditions could be implemented, e.g. following more recent works on ROA analysis which explored the use of invariant sets \cite{ValmorbidaAuto2017} or IQC-based formulation \cite{Iannelli_Auto_IQC}.
\end{remk}

\subsection{Identifying ODE using priors}
\label{sec:ode_estimation}
The purpose of this step is to estimate $f_{\bar{x}}$ (\ref{eq:dynamical_system_discrete_1}) from the observed trajectories (\ref{eq:dynamical_system_discrete_2}) using machine learning techniques, specifically neural networks~\cite{bishop:2006:PRML}.

It is well known that learning a complex nonlinear function from scarce data is prone to overfitting \cite{scholkopf2002learning}. A remedy, if the information on the functional form of the unknown function is available, consists of restricting the hypothesis space to the set of functions that is more likely to contain it. In this work, we do not make this assumption, and instead, propose to address this problem as a regularization strategy which leverages the ROA estimate $\mathcal{R}_{\bar{x}}$ computed in the previous section.
As it will be detailed in the rest of the section, the objective is to learn $f_{\bar{x}}$ by fitting the time derivatives of the observed trajectories with the vector field, while avoiding known pitfalls of learning approaches with a Lyapunov-type of regularization.

\subsubsection{Learning interpolants}
\label{sec:learning_interpolants}
Measurements typically give the states $x_t$ at some discrete times (\ref{eq:dynamical_system_discrete_2}), and this causes two issues. Firstly, the states might be only available on irregular (non-equidistant) and possibly sparse time intervals. Secondly, the noise in the state measurements will be magnified when the time derivative of $x_t$
is computed by methods such as finite-difference. To overcome both problems, a kernel interpolation method to artificially add data at intermediate times and also smooth the noisy observations is proposed.

The first step consists of fitting a scalar function $\hat{x}_s(\cdot;\phi_s);\RR\to\RR$, parametrized with $\phi_s$ and called~\emph{interpolant}, to the observed trajectories. More precisely, an interpolant is fitted to each component $s$ of the state vector, for each generated trajectory. This function interpolates between the times for which the states are measured and gives a continuous function from time to each dimension of the state vector. By choosing a differentiable kernel $k:\RR \times \RR\to \RR$, the interpolant is represented as:
\begin{equation}
        \hat{x}_s(t;\phi_s)=\sum_{i=1}^{m}\phi_{s,i}k(t, t_i),
\end{equation}
in terms of measured states at $m$ times. The choice of kernel for each state interpolant reflects our prior knowledge about that state. Radial Basis Function (RBF) are used here, which are defined as $k(a, b)=\exp(-\lVert a-b\rVert^2/2\sigma^2)$ where $\sigma$ captures how fast a state changes with time. Inspired by~\cite{gonzalez2014reproducing}, the following loss function is minimized to fit an interpolant to the $s^{\rm th}$ dimension of the noisy trajectory $y_{0:T}$
\begin{equation}
        \label{eq:interpolant_loss}
        \Lcal_\phi=\sum_{t=0}^T \frac{1}{2\sigma^2}|\hat{x}_s(t;\phi_s) - y_{s,t}|^2 + \frac{\lambda_\phi}{2}|\hat{x}_s(t)|^2
\end{equation}
Once $\hat{x}_s(t;\phi_s)$ is estimated for every trajectory and every dimension of the state vector, the rest of the dynamics learning algorithm is performed by evaluating these smooth functions to obtain the ground truth. In addition to addressing the problem of differentiating discrete variables by introduction of smoothness, it is worth observing that despite the fact that state trajectories are only available at discrete times, $\hat{x}_s(t;\phi_s)$ can be evaluated continuously. Let index $j$ iterates over the number of trajectories. Hence $\hat{x}_s^j(t;\phi)$ is the interpolant which is fit to the $s^{\rm th}$ dimension of the state vector of the $j^{\rm th}$ trajectory. A pseudocode regarding the computation of the interpolants is displayed in Algorithm~\ref{alg:interpolation}.

\subsubsection{Learning dynamics}
\label{sec:learning_dynamics_after_interpolants}
A method called~\emph{gradient matching} is used to fit $\hat{f}(\cdot;\psi)$ to interpolants $\hat{x}(\cdot;\phi)$. The idea is quite natural. It tries to fit $\hat{f}(\cdot;\psi)$  which is parameterized by a multilayer perceptron here to the time derivative of states which is computed by the analytical derivative of interpolants in our algorithm. Notice that interpolants have been already fitted to the observed trajectories as shown in Algorithm~\ref{alg:interpolation}.

The proposed loss function is
\begin{equation}
        \label{eq:gradient_matching}
        \Lcal'_{\psi}=\sum_{j=1s}^{J}\sum_{t=0}^T \lVert \differentiate{t}{\hat{x}^j(t;\phi)} - \hat{f}(\hat{x}^j(t;\phi);\psi)\rVert
\end{equation}
which is minimized with respect to $\psi$. Notice that $T$ is a typical length of the trajectory and can differ for each interpolant. $\hat{x}(t;\phi)$ is a compact notation for the vector $\hat{x}=[\hat{x}_1,\ldots,\hat{x}_n]$ of interpolants where $\phi=[\phi_1,\ldots, \phi_n]$. It can be observed that~\eqref{eq:gradient_matching} can be seen as a typical regression problem, i.e., the temporal sequence does not matter from the perspective of~\eqref{eq:gradient_matching}.

\begin{algorithm}[t!]
        \caption{Interpolation}
        \label{alg:interpolation}
        \SetKwInOut{Input}{input}\SetKwInOut{Output}{output}
        \Input{\begin{itemize}
                \item $\{y^j\}_{j=1}^J$: $J$ noisy state trajectories form $J$ different initial points
                \item kernel function $k(\cdot,\cdot)$
        \end{itemize}}

        \Output{$\Phi$ consisting of $\phi_s^j$ for all $s$ and $j$ }
        \For{$j=1,\ldots,J$}{
                \For{$s=1,\ldots,d$}{
                initialize $\phi$ in $\hat{x}(\cdot;\phi)=\sum_{i=1}^m\phi_i k(t,t_i)$\\
                compute $l_\lambda(\hat{x}|y_s^j;\sigma)$ as in~\eqref{eq:interpolant_loss}\\
                $\phi_s^j \leftarrow \argmin_{\phi} l_\lambda(\hat{x}|y_s^j;\sigma)$\\
            }
                    }
\end{algorithm}

\subsection{A coupled ODE-ROA algorithm}
\label{sec:learning_together_ode_roa}
The information of ROA is leveraged to learn the ODE in two ways: experiment design and regularization. Notice that the ROA is not known in advance. The algorithm proposed in Section~\ref{sec:roa_estimation} is a multi-step method that gradually expands the ROA. The ODE learning method is interlaced with the steps of the ROA estimation algorithm. Hence, the algorithm gradually improves its estimate of the ROA and estimates the ODE restricted to its current estimate of the ROA. The iterative process continues until the ROA cannot be improved any further. The estimated ODE will then be an approximation of the true dynamics within the ROA associated with the considered equilibrium.

\subsubsection{Experiment design} In this phase, the ROA information is used to choose initial states from which the trajectories are produced. Since the objective is to estimate $f_{\bar{x}}$ within the ROA, this method is able to choose initial states (experiments) that would save the computational time (or experimental costs) for the experiments that do not contribute to the accuracy of the estimated ODE within the ROA. Assume $\Vcal_c$ is the current estimate of ROA which is associated with the level set $V(x;\theta)=c$ of the current estimate of the Lyapunov function. Then new initial states are chosen randomly from the gap region $\Gcal=\Vcal_{\alpha c}\backslash\Vcal_\alpha$ whose width is determined by $\alpha>1$. We observed it is useful to keep the trajectories produced by the chosen initial states over the iterations of the algorithm. This prevents the algorithm from the so-called~\emph{catastrophic forgetting} where some previously acquired knowledge about the ODE is lost once the parameters of the neural network are updated by the trajectories chosen from the new gap region. Hence, the dataset to learn the ODE contains all the trajectories produced so far. However, trajectories belonging to the previous iterations of the algorithm are down-weighted by the factor $\epsilon^{\Delta}$ where $\Delta$ is the difference between the current iteration and the iteration of the algorithm in which a trajectory has been produced. A value $\epsilon=0.8$ was observed to work well for a range of experiments.

\subsubsection{Regularization}
In the regularization phase, the Lyapunov function whose level set is the current estimate of the ROA is employed directly. The ODE must comply with the decrease condition of the Lyapunov function within the ROA (according to Lemma \ref{lemma:LF_LevelSets}). This structural constraint of the identification is incorporated in the loss function as a Lagrange multiplier. Therefore,~\eqref{eq:gradient_matching} gets augmented as the following
\begin{equation}
        \Lcal_\psi = \Lcal'_\psi + \lambda_\psi \sum_{x\in{\rm \Vcal_c}}\langle\nabla_x V(x;\theta), \hat{f}(x;\psi)\rangle
        \label{eq:Lyapunov_regularizer}
\end{equation}
which is to be minimized with respect to $\psi$ for a fixed $\theta$. This term encourages learning dynamics $\hat{f}$ for which the learned Lyapunov function attains negative derivatives.

A pseudocode of the co-learning algorithm is reported in Algorithm~\ref{alg:ours}, and is briefly summarized in the following. The ROA is expanded in multiple stages as described in~\ref{sec:roa_estimation}. At every stage, the ROA information is used to choose the initial states to produce trajectories required for learning the ODE. Moreover, the associated Lyapunov function is used to regularize the learned dynamics functions and by doing so estimations featuring this local stability constraint are favored.

\begin{algorithm}[t!]
        \caption{\ours}
        \label{alg:ours}
        \SetKwInOut{Input}{input}\SetKwInOut{Output}{output}
        \Input{\begin{itemize}
                \item $f$, The oracle that generates trajectories from the system
                \item $\Vcal_{\rm init}$, Initial inner estimate of the ROA
                \item $T$, length of the trajectories
        \end{itemize}}
        \Output{\begin{itemize}
                \item $V(\cdot;\theta), c$, The Lyapunov function $V$ and its level set $c$ that estimates the true ROA
                \item $f(\cdot;\psi)$, The estimates of the dynamics function
        \end{itemize}}
        init $V(\cdot;\theta)$ to the initial inner estimate of ROA\\
        init $\psi$ by a zero-mean Gaussian distribution with standard deviation of $0.1$\\
        set the interpolant kernel function $k(\cdot, \cdot)$\\
        set the growth coefficient $\alpha$\\
        set the growth threshold $\Tcal_g$\\
        set ${\rm Interpolants} \leftarrow [\ ]$\\
        Set $\Gcal\leftarrow \Vcal_{\alpha c} \backslash \Vcal_c$\\
        \While{$\vol(\Gcal) > \Tcal_g$}{
                $\{x^j\}_{j=1}^J \leftarrow$ Sample $J$ initial points from $\Gcal$\\
                $\{\tau^j\}_{j=1}^J \leftarrow$ Tarjectories starting from $\{x^j\}_{j=1}^J$\\
                $\Phi \leftarrow$ compute interpolants from $(\{\tau^j\}_{j=1}^J, k)$ (by Algorithm~\ref{alg:interpolation})\\
                ${\rm Interpolants}\leftarrow {\rm Interpolants } \cup \{\hat{x}^j(\cdot;\Phi^j)\}_{j=1}^J $\\
                $\psi\leftarrow$ Update ODE $({\rm Interpolants}, \theta)$ (by~\eqref{eq:Lyapunov_regularizer})\\
                $\theta, c\leftarrow$ Expand ROA $(\theta, \Gcal)$(by Section~\ref{sec:roa_estimation})\\
                $\Gcal\leftarrow \Vcal_{\alpha c}(\theta)\backslash\Vcal_c(\theta)$
            }
            
\end{algorithm}

\begin{figure}[t!]
        \centering
        \subfigure[\scriptsize Random network]{\includegraphics[width=0.3\linewidth]{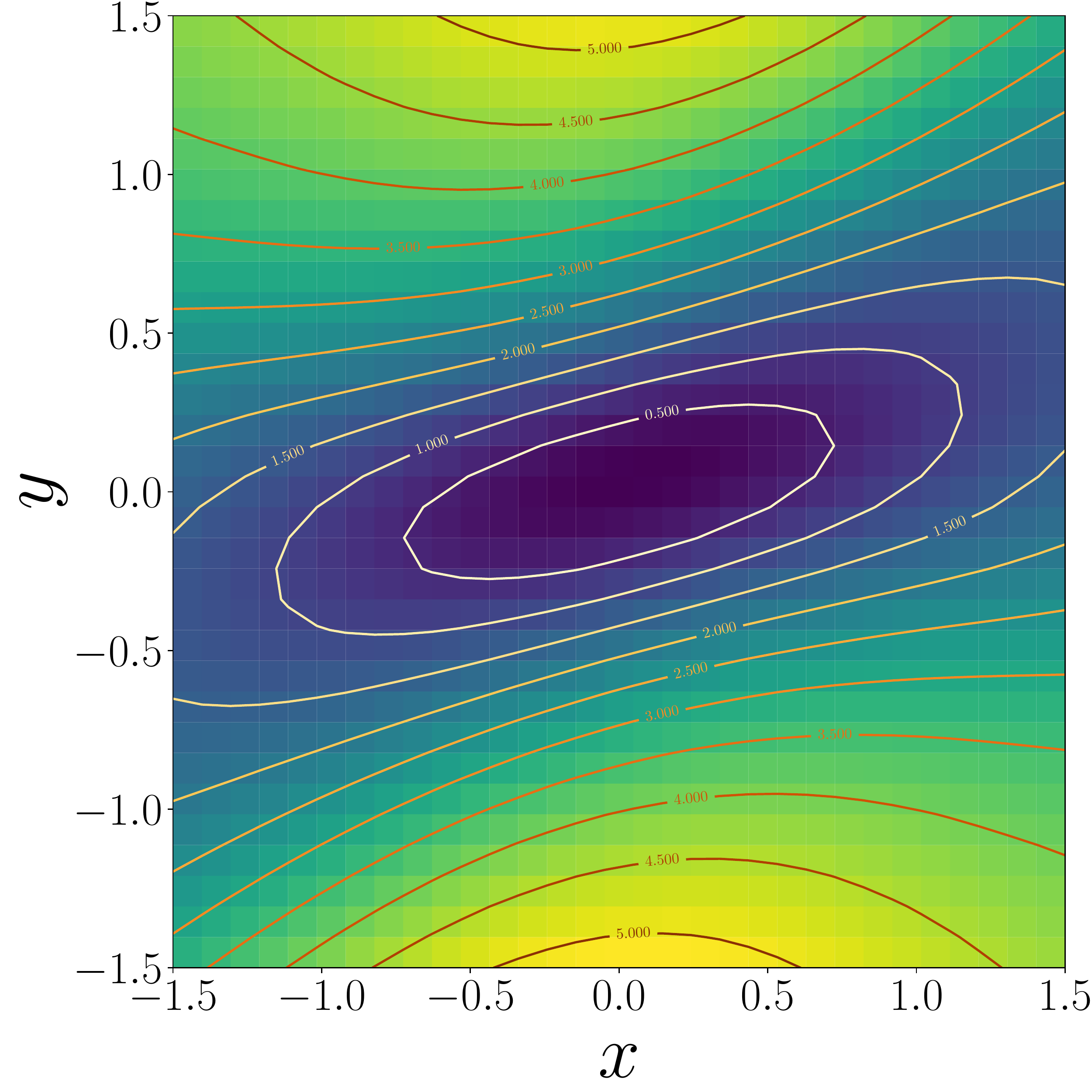}}
        \subfigure[\scriptsize Quadratic function]{\includegraphics[width=0.3\linewidth]{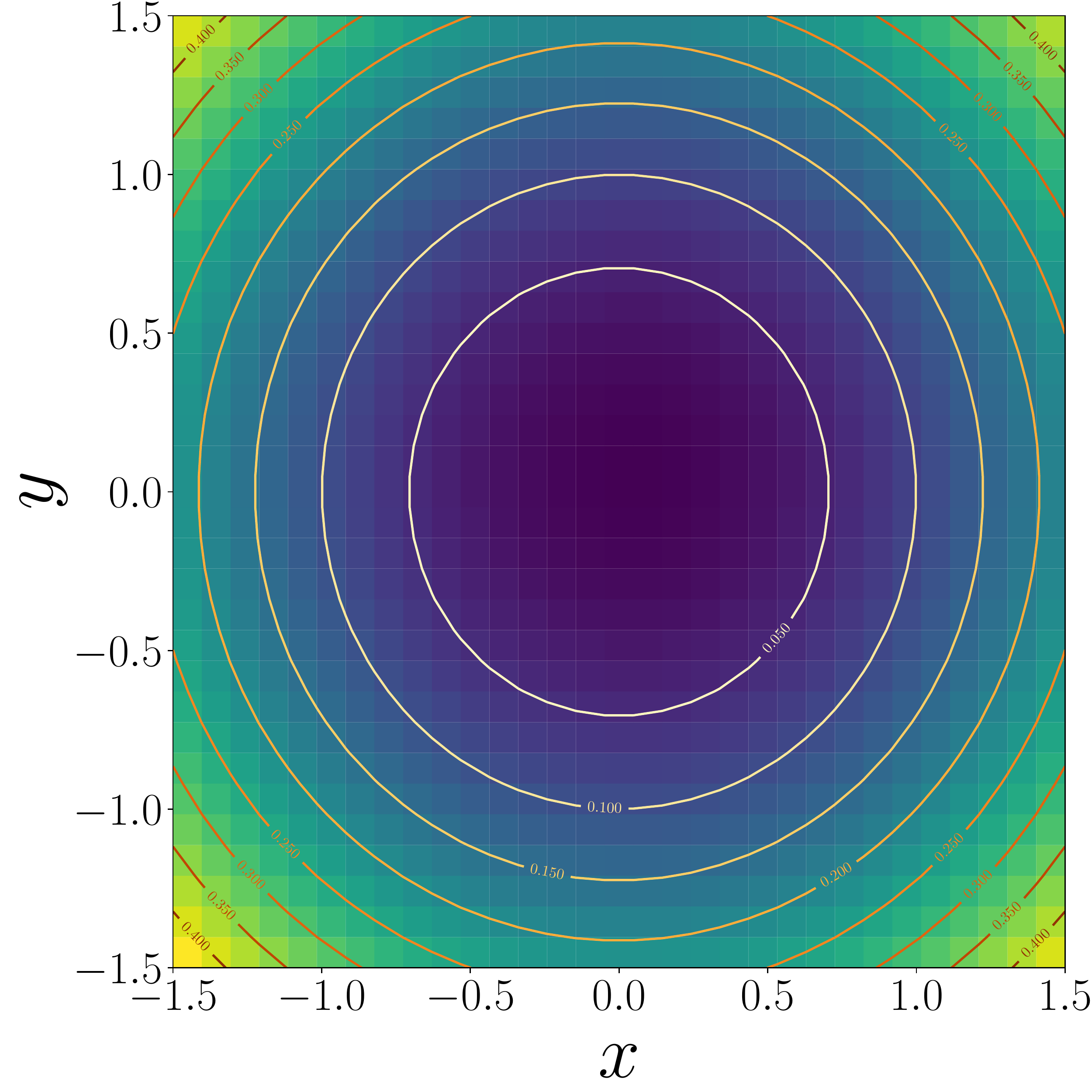}}
        \subfigure[\scriptsize Pretrained network]{\includegraphics[width=0.3\linewidth]{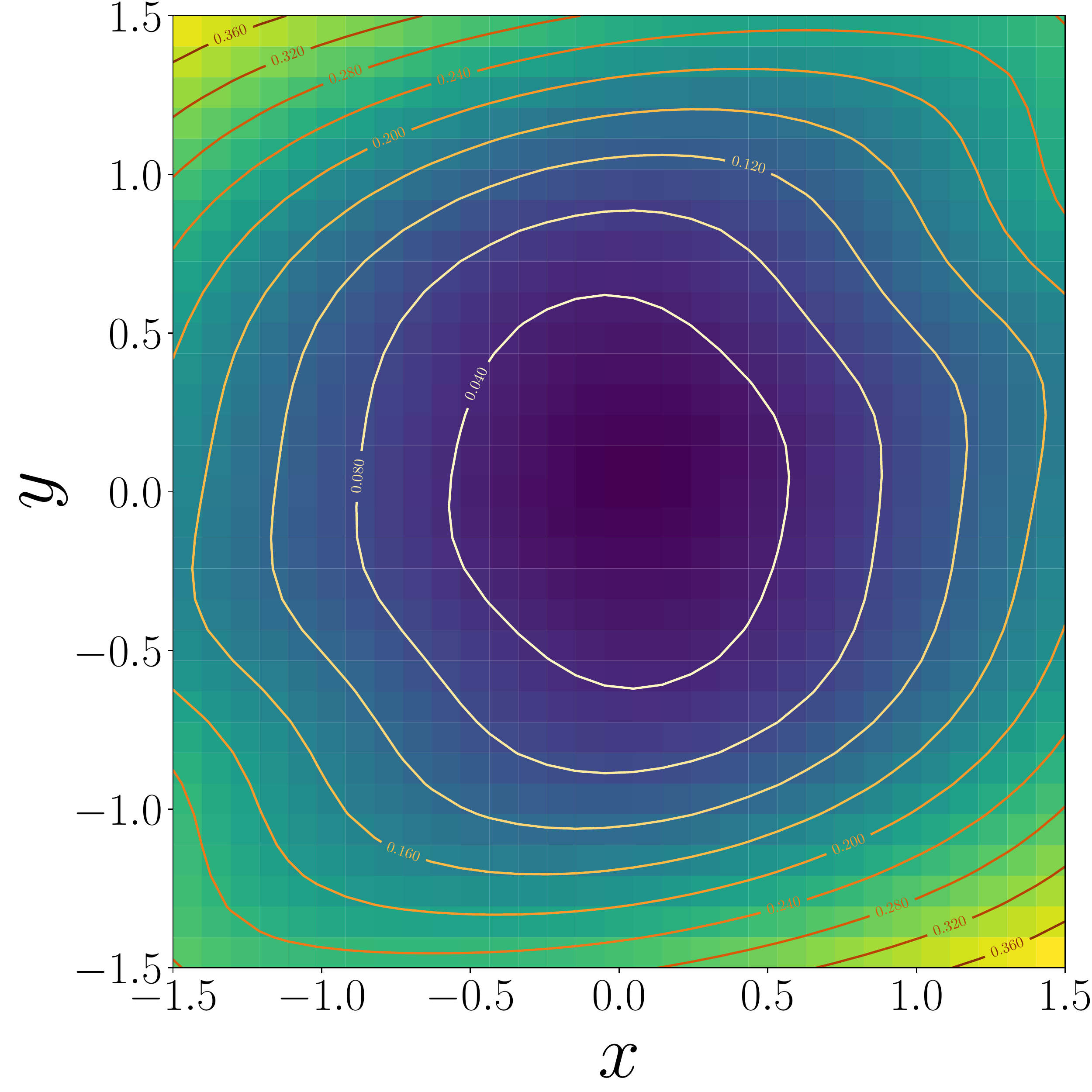}}
        \caption{Pre-training the randomly initialized neural network with a quadratic function. The background color shows the values of the function from $\RR^2$ to $\RR_+$. Lighter colors correspond to larger values. The contours show the levelsets.}
        \label{fig:roa_pretraining}
\end{figure}

\begin{figure}[t!]
        \centering
        \subfigure[\scriptsize Growth Stage 0]{\includegraphics[width=0.45\linewidth]{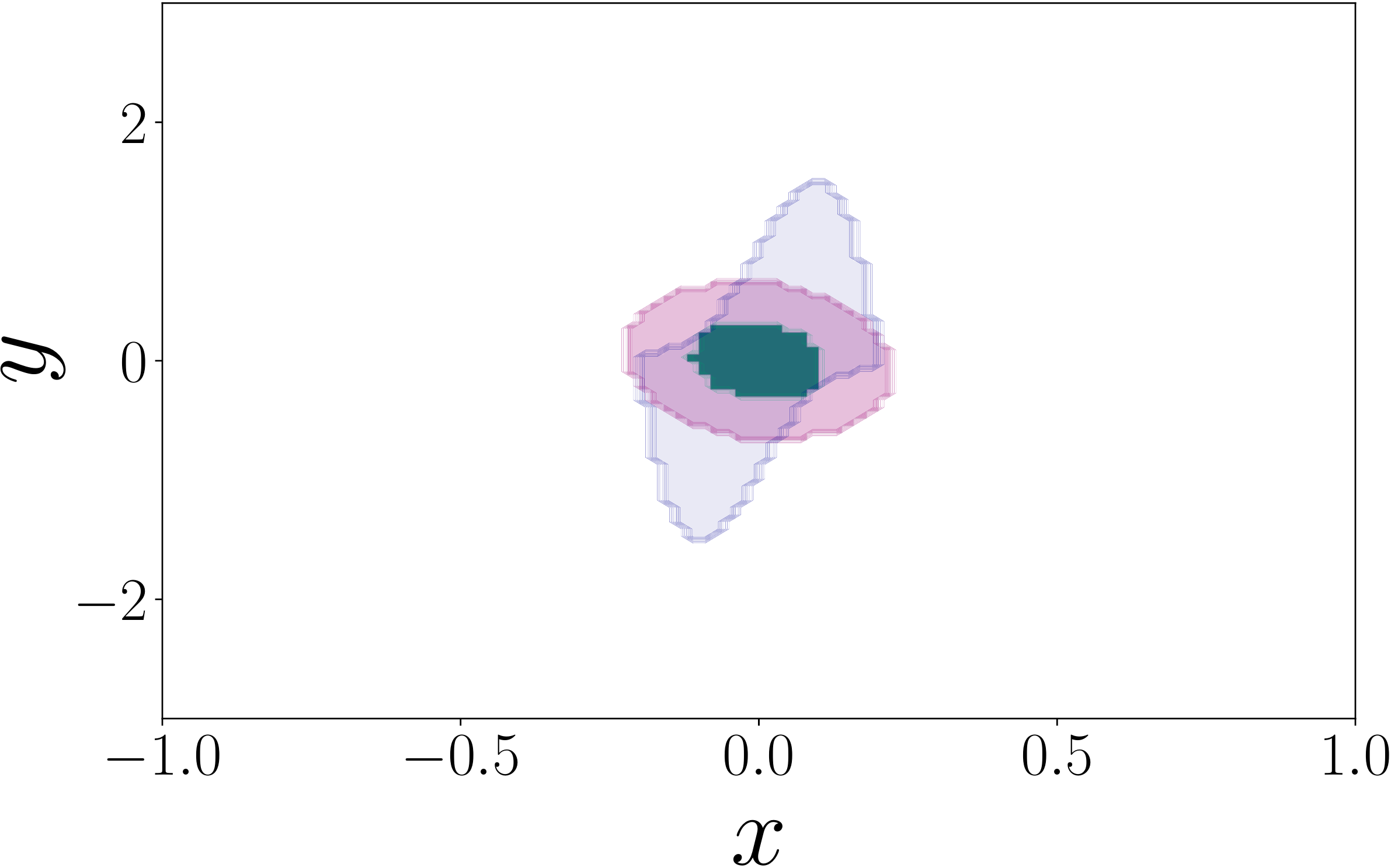}}
        \subfigure[\scriptsize Growth Stage 1]{\includegraphics[width=0.45\linewidth]{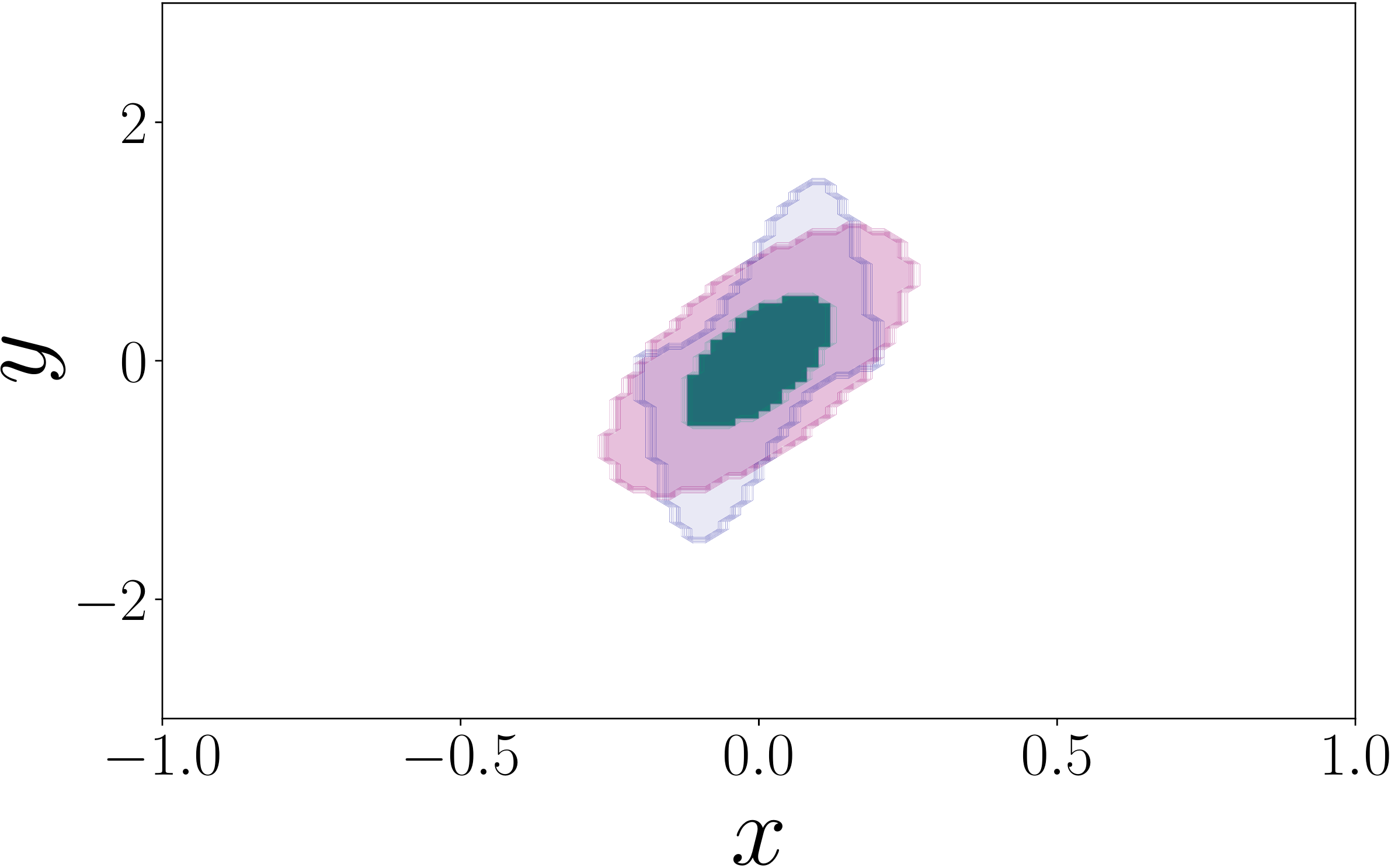}}
        \subfigure[\scriptsize Growth Stage 4]{\includegraphics[width=0.45\linewidth]{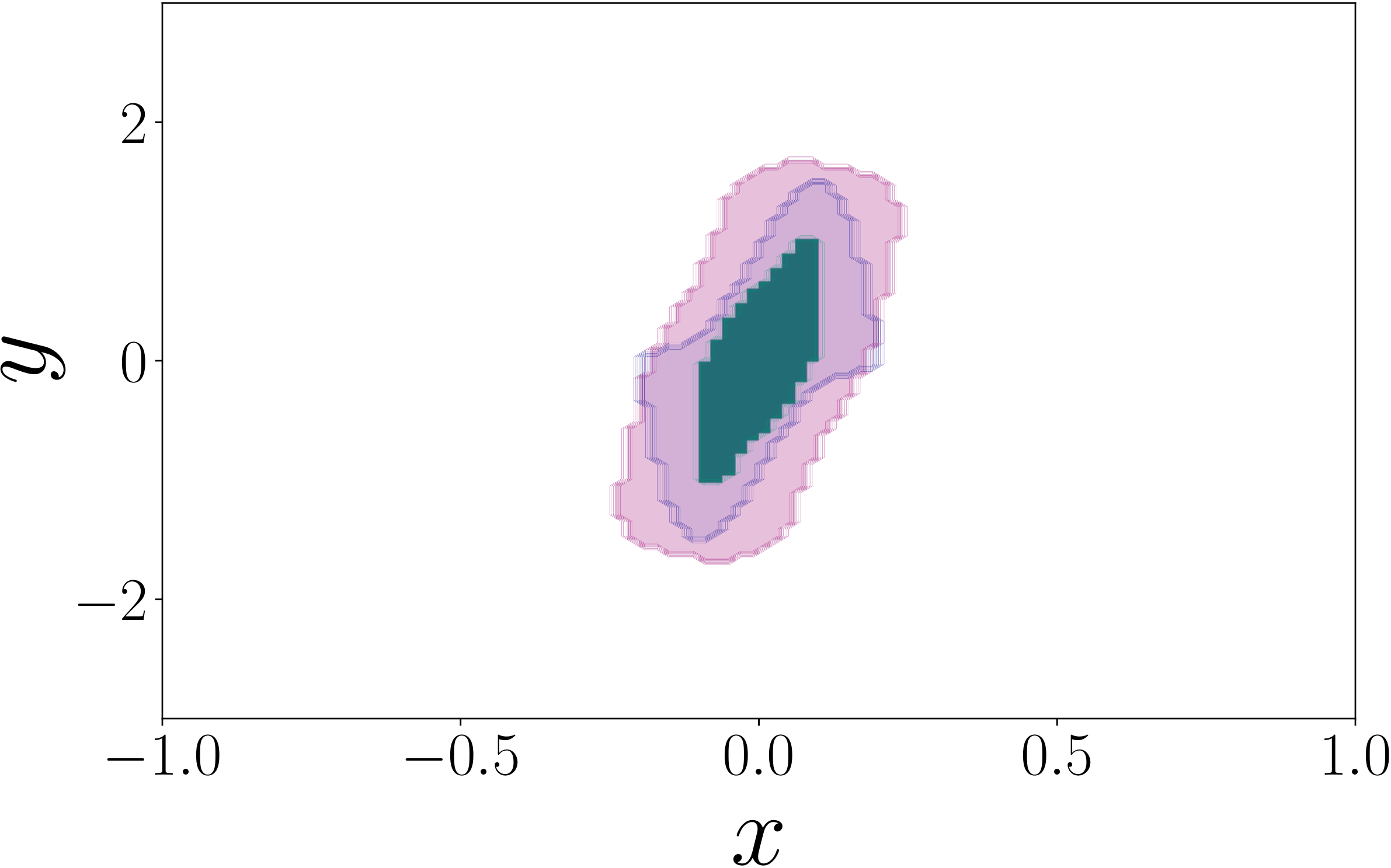}}
        \subfigure[\scriptsize Growth Stage 6]{\includegraphics[width=0.45\linewidth]{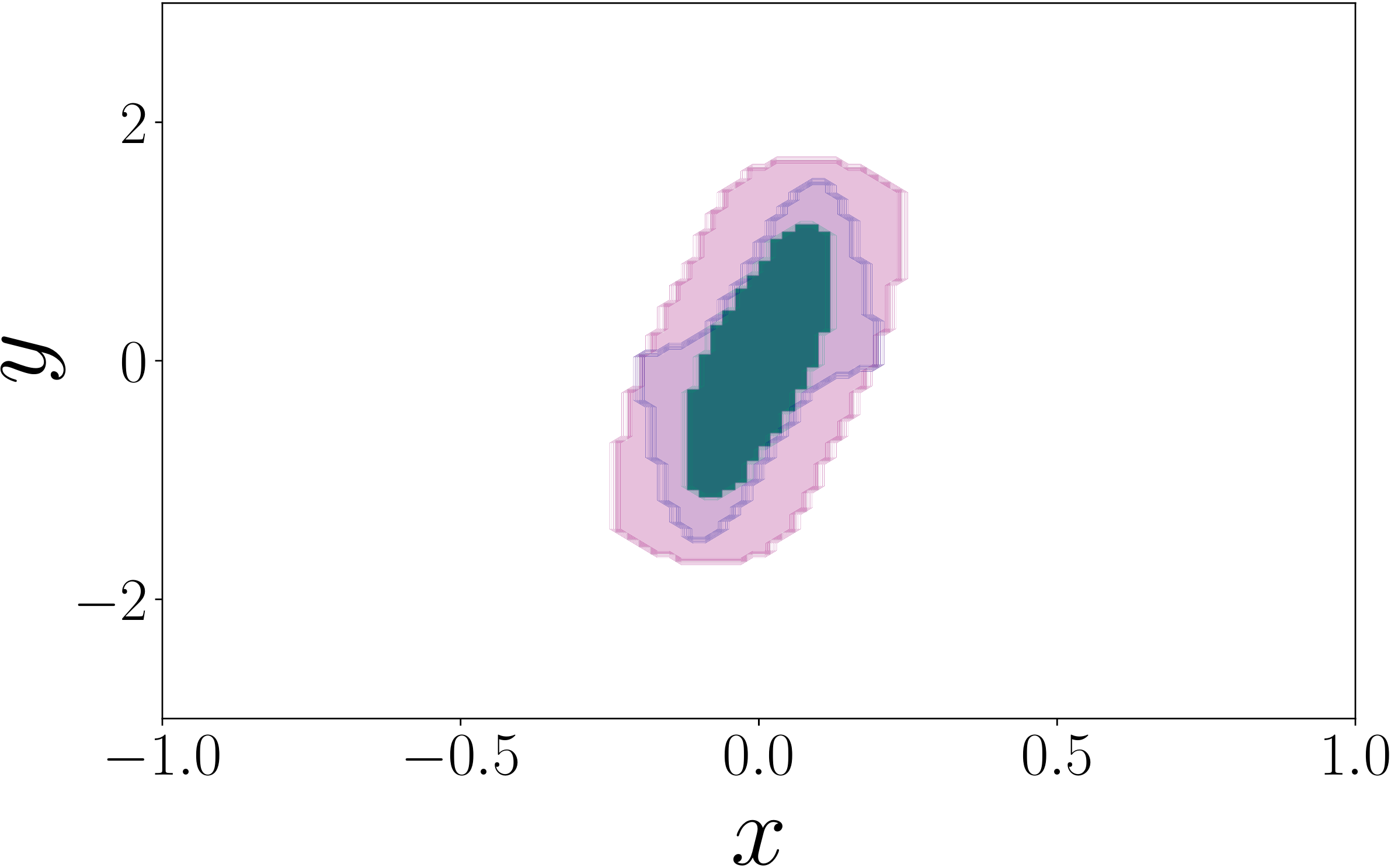}}

        \caption{The growth stages of the ROA estimation algorithm. The scale of axes is changed for better visibility. The meaning of the colors is as follows. Green: The estimated ROA corresponding to $\Vcal_c$. Pink: The gap $\Gcal=\Vcal_{\alpha c}\backslash \Vcal_c$. The blue contour: The boundary of the true ROA $\Rcal$.}

        \label{fig:nested_roas}
        \vspace{-2ex}
\end{figure}

\section{Numerical examples}
\label{sec:experiments}
In this section, the proposed framework for co-learning ROA and ODE is demonstrated on a well-known benchmark test case in the ROA analysis literature \cite{ValmorbidaAuto2017,TOPCU20082669,Chesi_book,Iannelli_Auto_IQC}.
The Van der Pol system is a $2-$dimensional dynamical system defined as:
\begin{equation}
        \label{eq:vanderpol}
        \begin{aligned}
        & \dot{x} = -y\\
        & \dot{y} = x + \gamma (x^2 - 1) y\\
        \end{aligned}
\end{equation}
where $\gamma$ is the damping parameter. When $\gamma>0$, the system has an unstable limit cycle around the equilibrium and the true ROA of the stable equilibrium (at the origin) is the area encircled by the limit cycle (which can be obtained by simulating Eq. \ref{eq:vanderpol} backwards in time). The value $\gamma=3$ is chosen here. All measured trajectories are contaminated with Gaussian noise with a standard deviation $0.05$ (recall the measurement equation ~\eqref{eq:dynamical_system_discrete_2}).

The focus is on the performance of the ROA estimation algorithm (described in Section~\ref{sec:roa_estimation}). It is recalled here that the Lyapunov function is parameterized as $V(\cdot;\theta) = v\tran(\cdot;\theta)v(\cdot;\theta)$, where $v(\cdot;\theta)$ is a neural network. The initial values of its weights $\theta_{\rm init}$ are i.i.d. samples from a zero-mean Gaussian distribution with a standard deviation of $0.1$. The corresponding initial shape of $\Vcal_c(\theta_{\rm init})$ can be far from the shape of the true ROA resulting in a negative impact on the learning algorithm. Numerical tests suggest for example that starting with a steep function (i.e. with a function $V(\cdot;\theta_{\rm init})$ with high-gradient) increases the chance to get stuck in the early growth stages of the algorithm and therefore obtain a highly conservative estimate of the ROA.

To address this aspect, it is proposed to pretrain the neural network with the loss function~\eqref{eq:pretraining_quadratic_loss} and results are shown in Figure~\ref{fig:roa_pretraining}. Figure~\ref{fig:roa_pretraining}(a) shows the level sets of the randomly initialized network (i.e. with no pretraining). After training with the loss function~\eqref{eq:pretraining_quadratic_loss} for $Q=I$, the level sets of the trained network (Figure~\ref{fig:roa_pretraining}(c)) become closer to those of the target quadratic function (Figure~\ref{fig:roa_pretraining}(b)).
We observed in all the tests that pretraining increases the stability of the algorithm significantly and results in a less conservative final estimate of the ROA.

After pretraining the network, the algorithm for learning ROA is run. The growth parameter $\alpha$ is set to $3$. Six growth stages of the algorithm are shown in Figure~\ref{fig:nested_roas}. The blue contour shows the boundary of the true ROA $\Rcal$, which is readily available from the simulation as said earlier. The dark green area represents the level set $\Vcal_c(\theta)$ computed at each growth stage by the algorithm. The pink shape is $\Vcal_{\alpha c}(\theta)$ and is interesting to observe because samples for learning the ROA are drawn at each stage from the gap region $\Gcal=\Vcal_{\alpha c} \backslash \Vcal_{c}$.
As can be seen, the stages start with the initialized network where the estimated ROA is similar to that of a quadratic function that lives within $\Rcal$. As the growing process progresses, $\Vcal_c(\theta)$ gets closer to $\Rcal$ in shape and size and always remains within it. 

\begin{figure}[t!]
        \centering
        \subfigure[\scriptsize Growth Stage 1]{\includegraphics[width=0.45\linewidth]{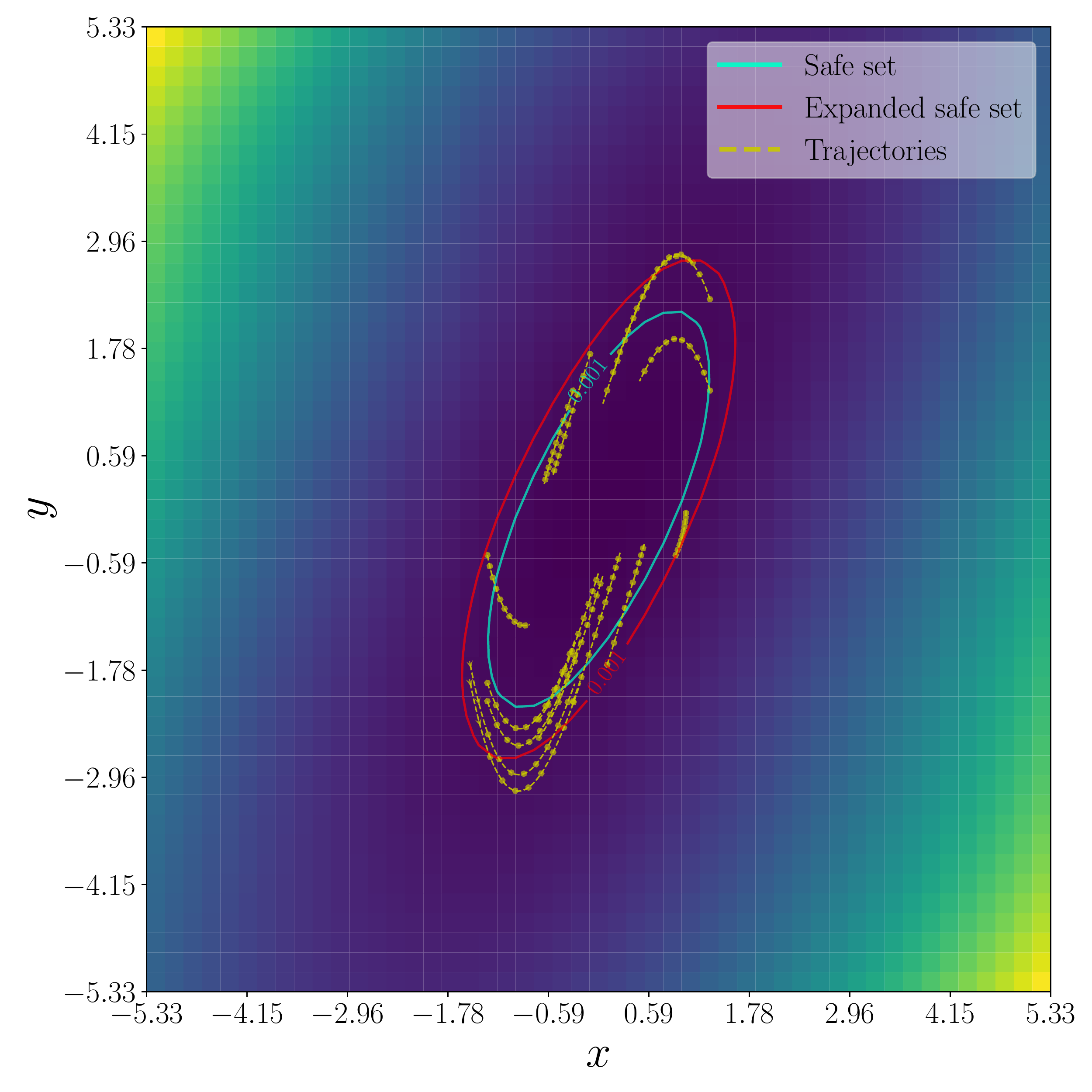}}
        \subfigure[\scriptsize Growth Stage 2]{\includegraphics[width=0.45\linewidth]{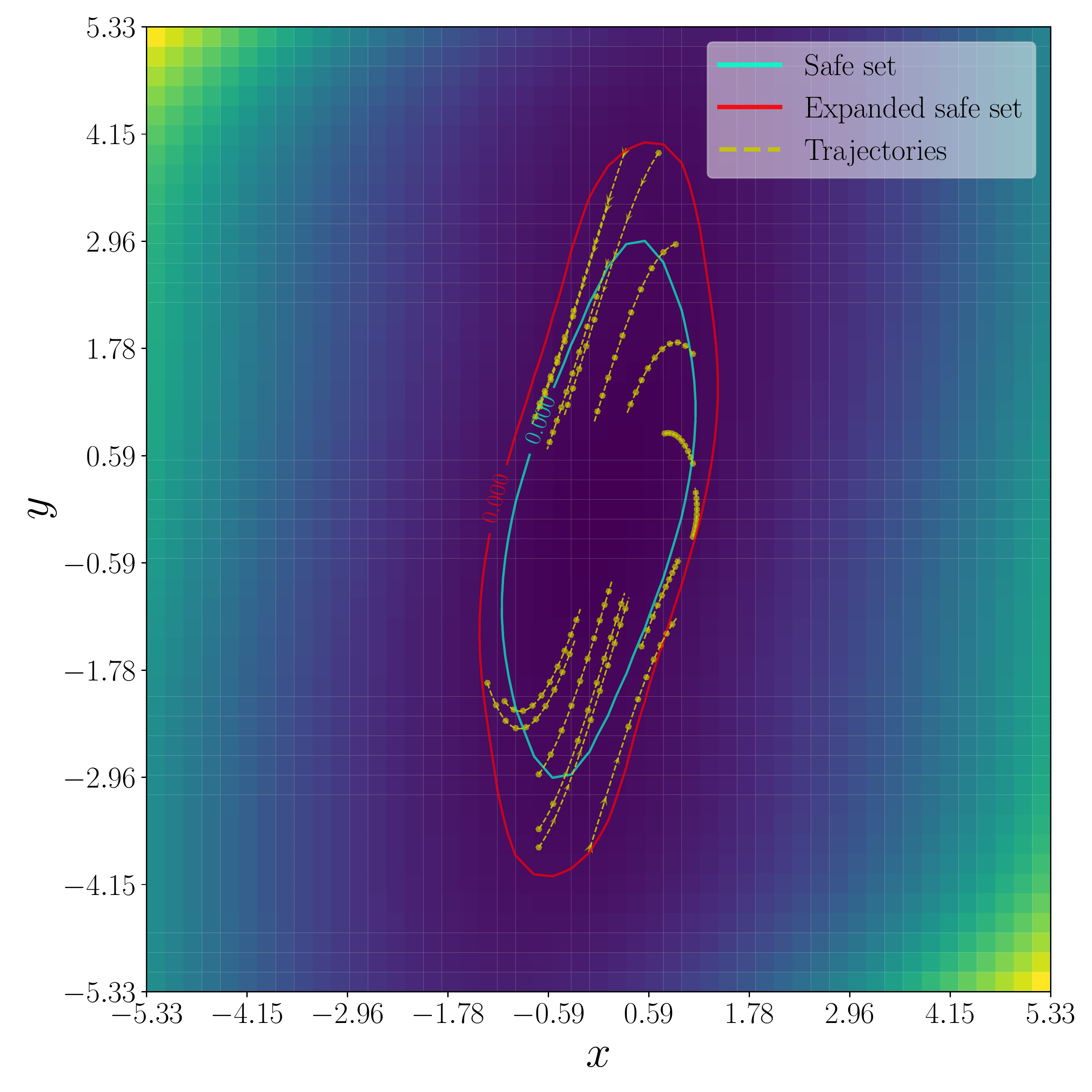}}
        \subfigure[\scriptsize Growth Stage 3]{\includegraphics[width=0.45\linewidth]{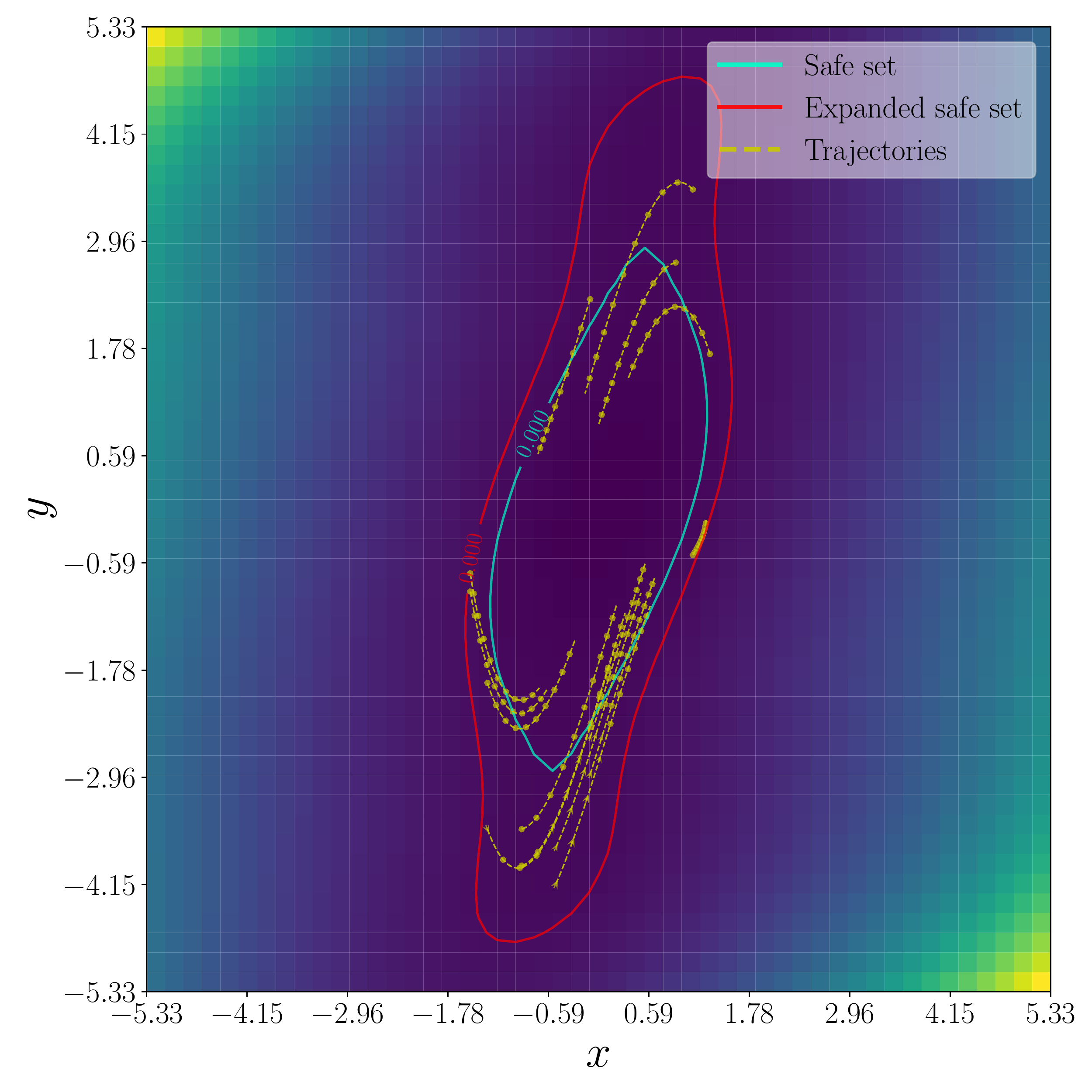}}
        \subfigure[\scriptsize Growth Stage 4]{\includegraphics[width=0.45\linewidth]{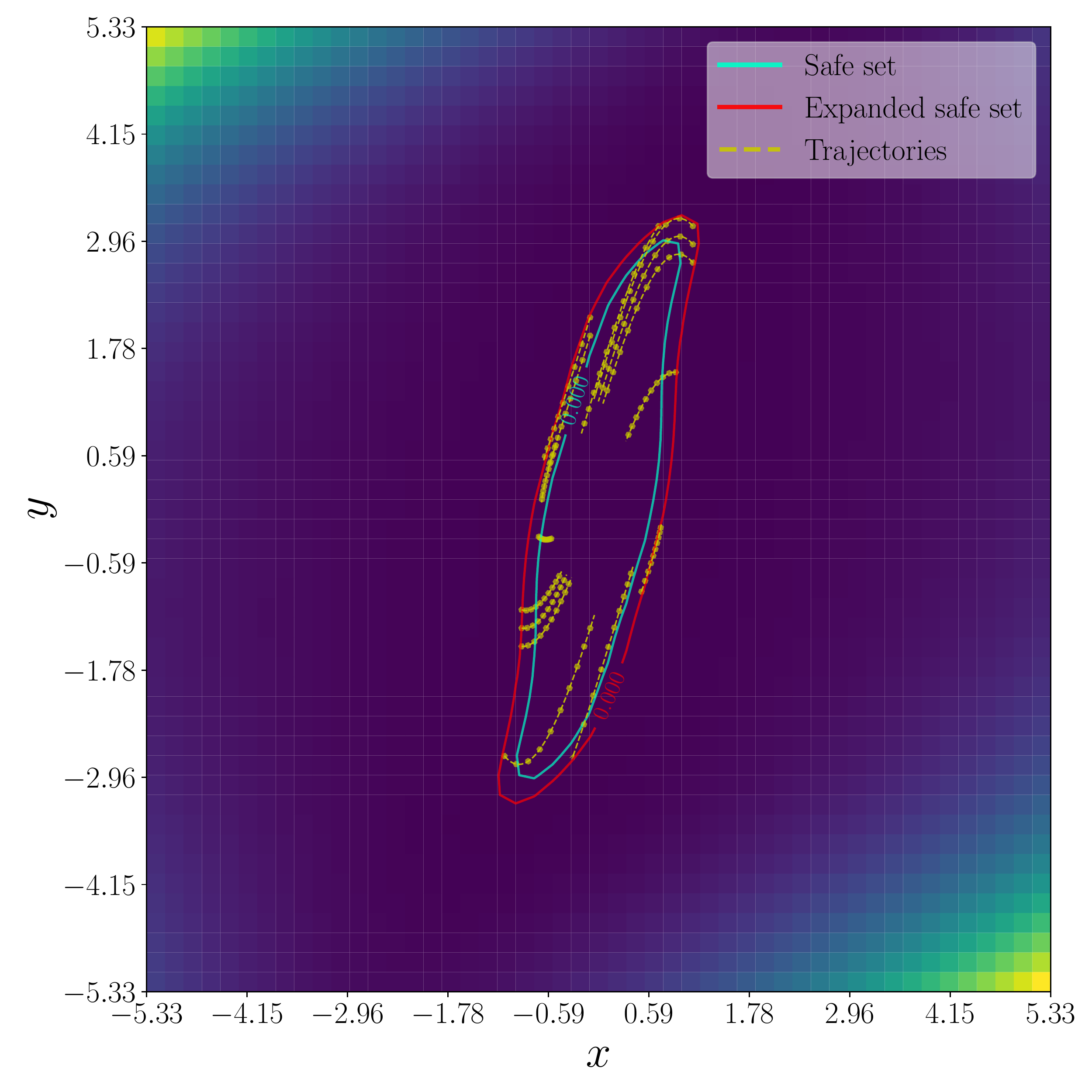}}
        \caption{Sampled trajectories from around the estimated ROA of each growth stage. The background color shows the values of a function from $\RR^2$ to $\RR_+$. Lighter colors correspond to larger values.}.
        \label{fig:trajs_on_levelsets}
\end{figure}

\begin{figure}[t!]
        \centering
        \subfigure[\scriptsize Growth Stage 0]{\includegraphics[width=0.45\linewidth]{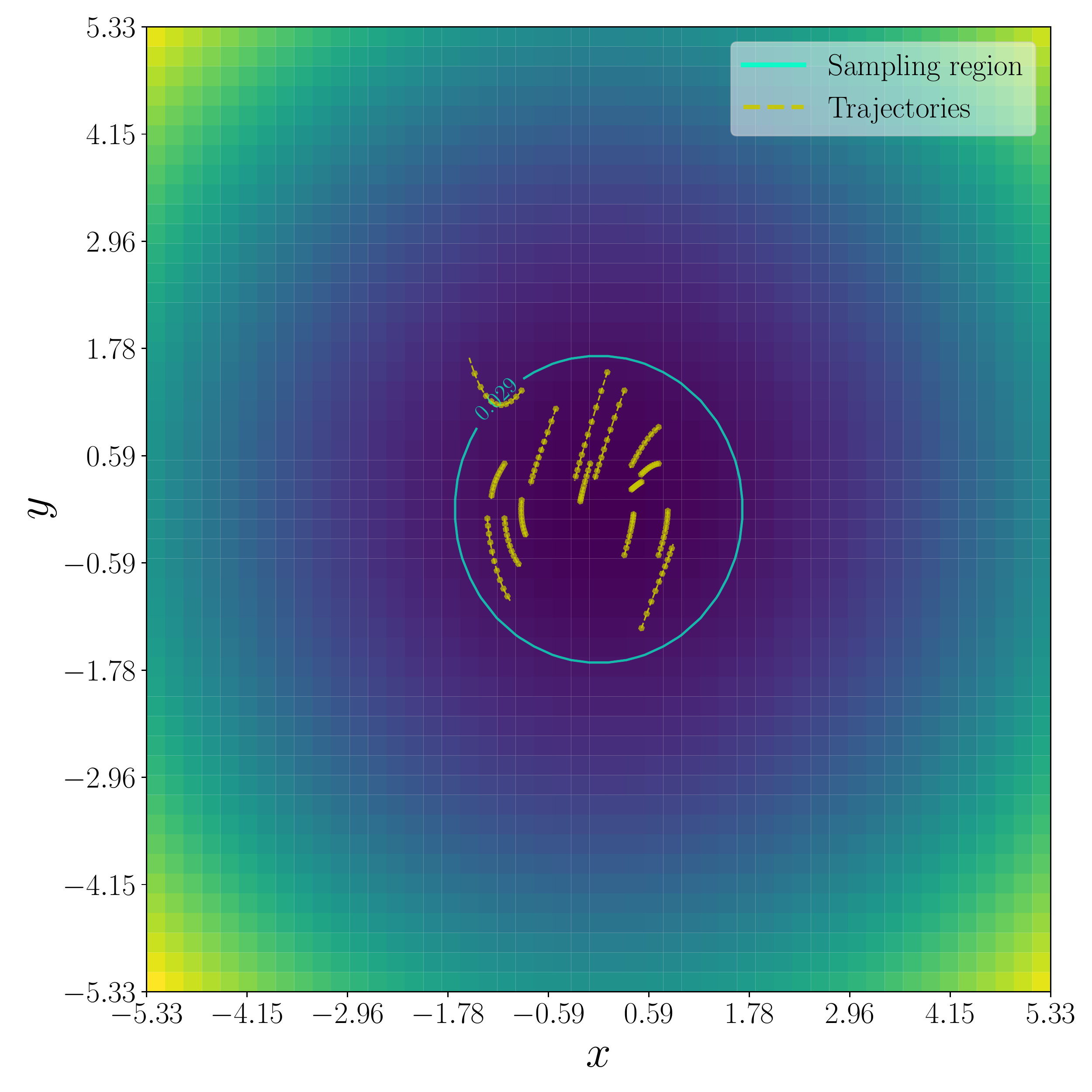}}
        \subfigure[\scriptsize Growth Stage 3]{\includegraphics[width=0.45\linewidth]{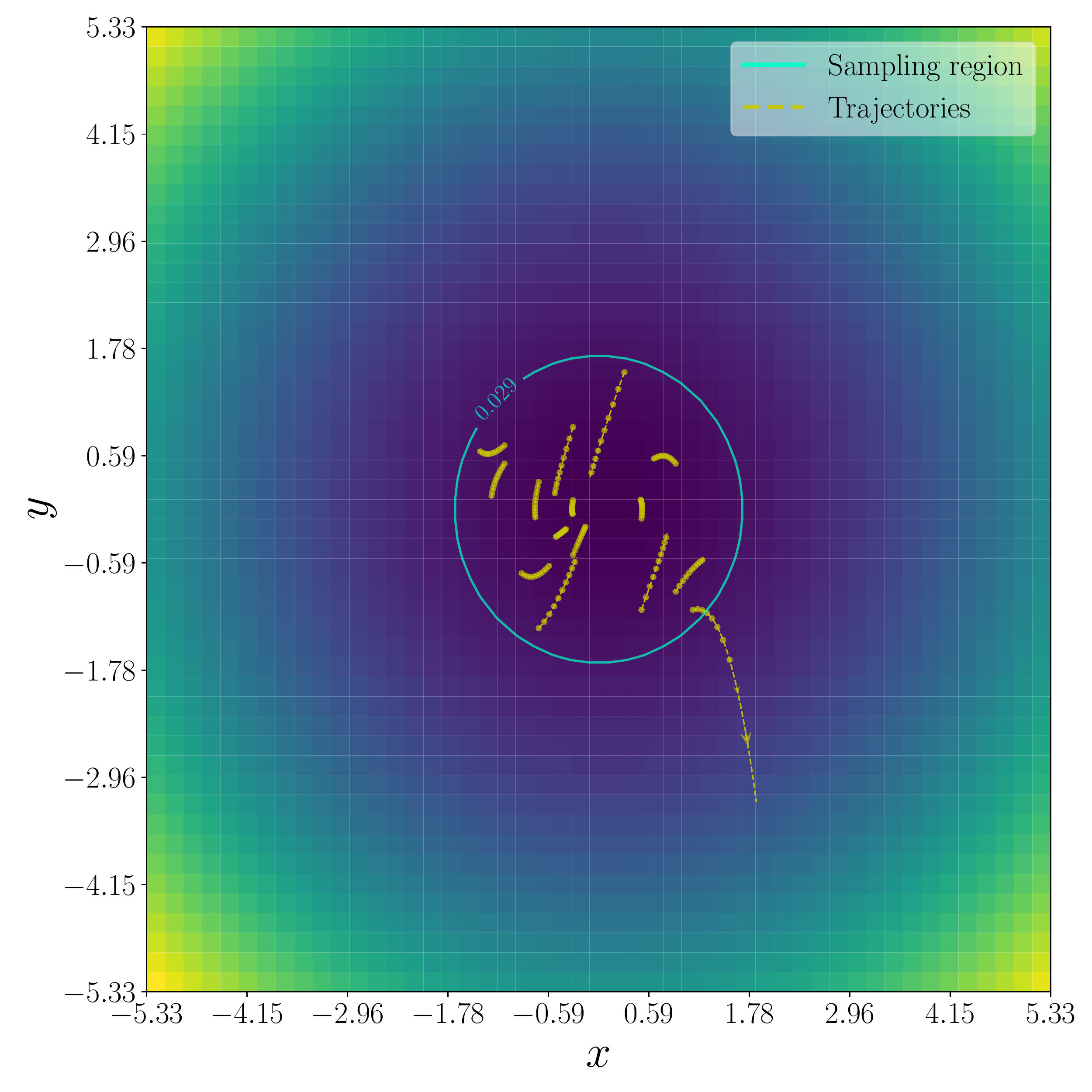}}
        \caption{Sampled trajectories from a ball around the equilibrium. The background color shows the values of a function from $\RR^2$ to $\RR_+$. Lighter colors correspond to larger values.}
        \label{fig:trajs_on_levelsets_blind}
\end{figure}

\begin{figure}[t!]
        \centering
        \subfigure[\scriptsize Ball sampling, Growth stage 5]{\includegraphics[width=0.3\linewidth]{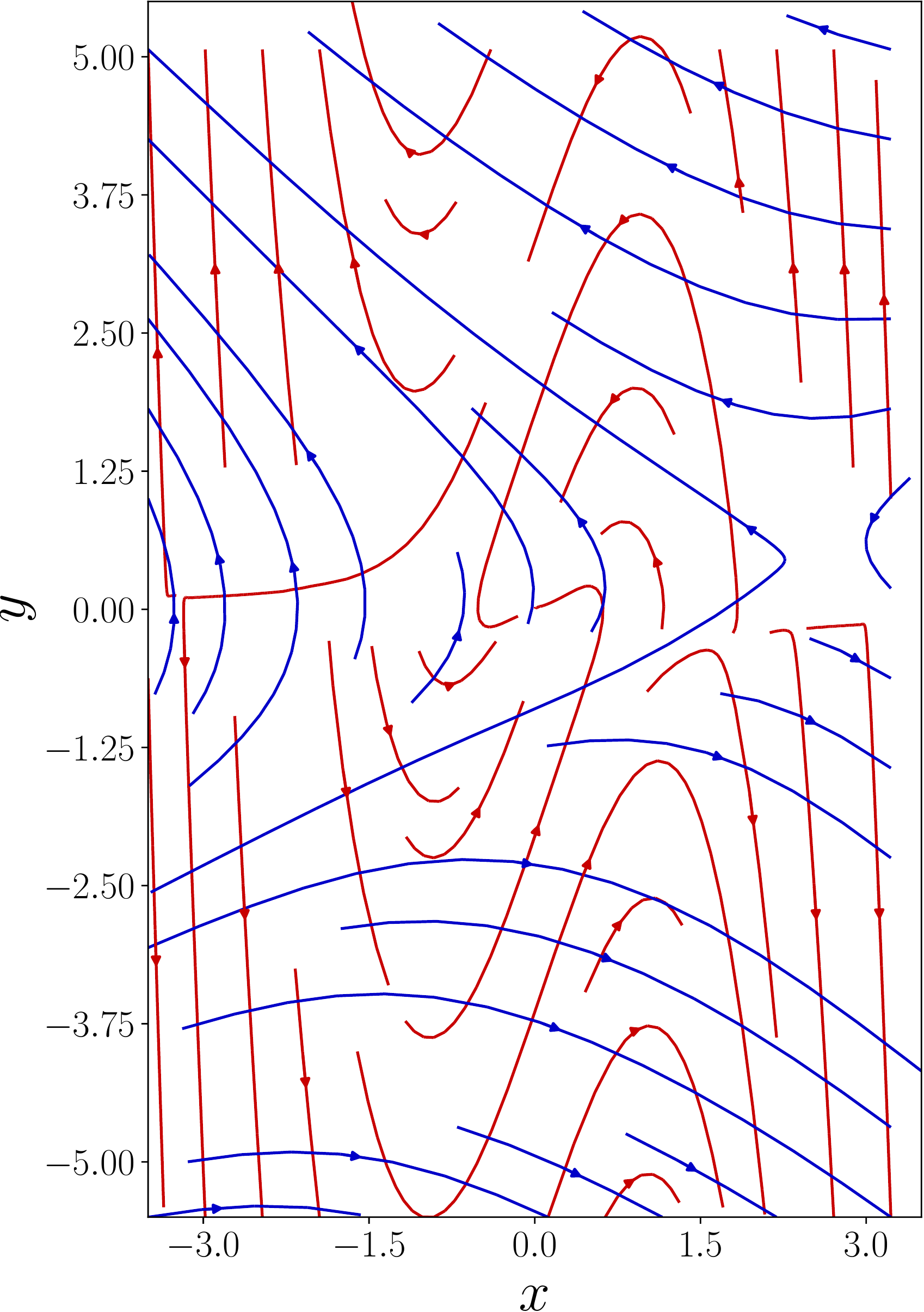}}
        \subfigure[\scriptsize ROA sampling, Growth stage 5]{\includegraphics[width=0.3\linewidth]{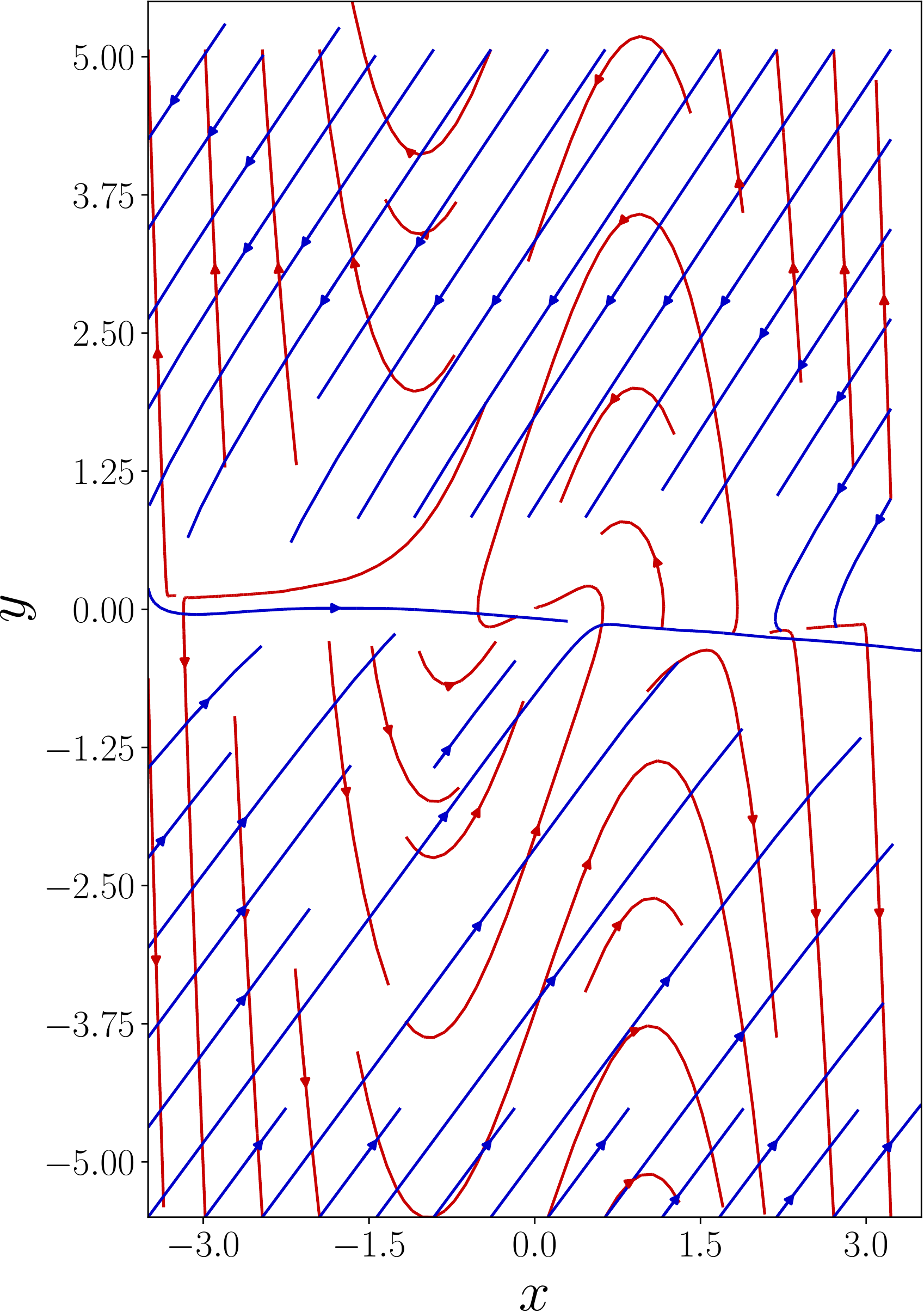}}
        \subfigure[\scriptsize ROA sampling + Regularizer, Growth stage 5]{\includegraphics[width=0.3\linewidth]{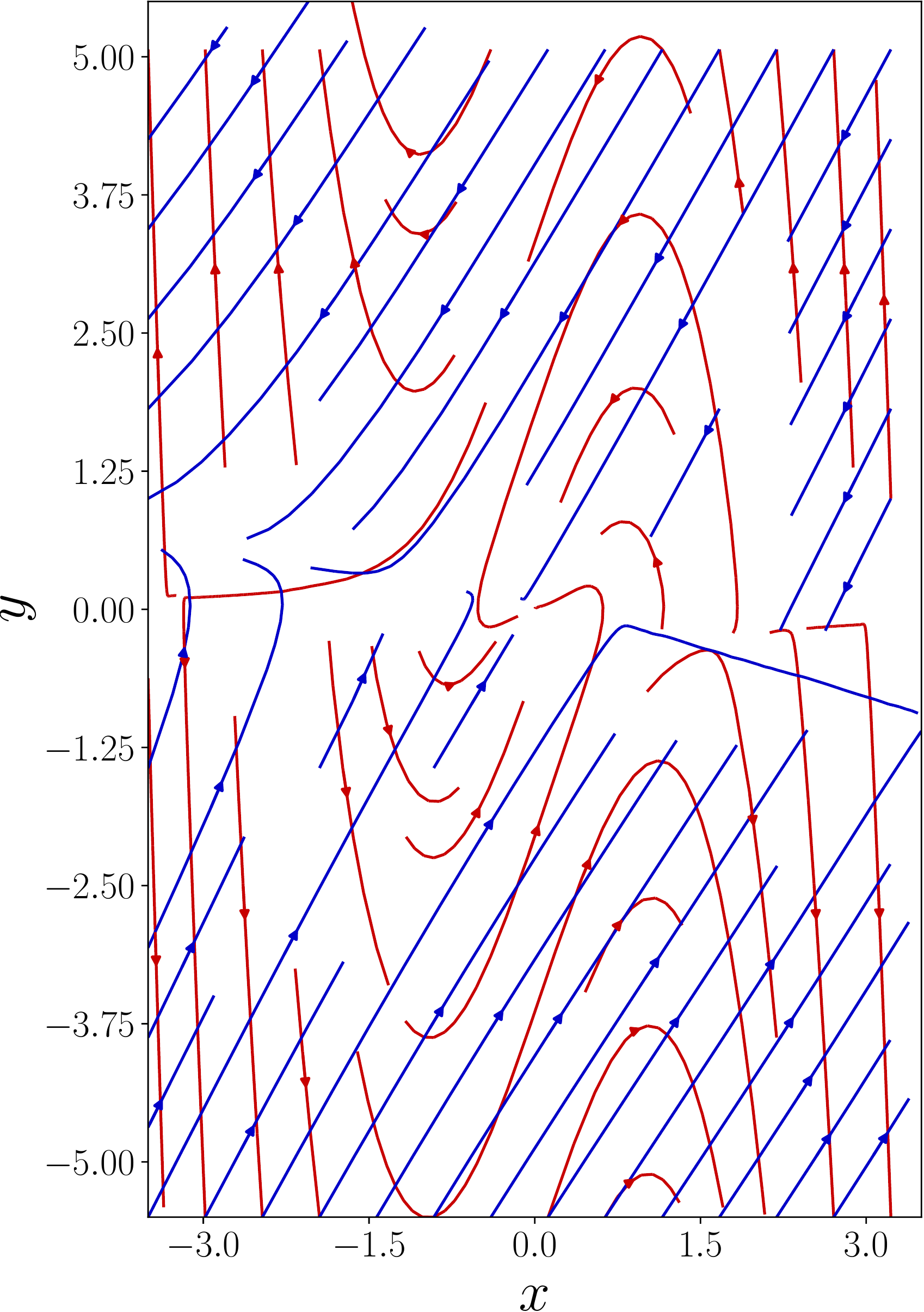}}

        \caption{The comparison of different ODE learning algorithms. Notice that the region of interest is the central area that locates within the ROA. Red: The true vector field. Blue: The learned vector field.}
        \label{fig:vector_fields}
\end{figure}

The iterations of \ours method (Algorithm \ref{alg:ours}) consists of expanding the ROA and sampling initial conditions near its boundary to generate trajectories for learning the ODE. Some of these trajectories are shown in Figure~\ref{fig:trajs_on_levelsets}. As can be seen, because the initial states are chosen using the information of the estimated ROA up to that growth stage, most of the sampled trajectories are stable and move towards the region of attraction. Hence, they contain more information about the ODE restricted to the ROA compared to the trajectories that are randomly sampled from the state space or from a specified set that is blind to the ROA information.
This latter instance is exemplified in Figure~\ref{fig:trajs_on_levelsets_blind}, where initial points are chosen by uniformly sampling from a ball around the equilibrium with radius $1$. Since the ball is chosen without having knowledge of the ROA, it can be seen that some of the chosen initial states result in unstable trajectories. Unstable trajectories have clearly a potentially harmful effect in an experimental setting, but they can also hamper the learning procedure by providing non-informative data as it will be later shown in Figure~\ref{fig:vector_fields}(a). The method proposed in this paper inherently provides a solution to this issue. Indeed, trajectories generated with the experiment design procedure described earlier are in an invariant set and thus will remain within it.

The trajectories are stored during the growth stages and used for learning the ODE with more weights on the newer trajectories. This is done to make sure the learned ODE becomes accurate in the newly expanded region around the current ROA while it does not forget the knowledge it has acquired in previous stages. As a result, the learned ODE at each growth stage will be equally accurate within the ROA associated with that stage.

\begin{table}[]
        \caption{Number of sampled trajectories and mean squared estimation error for three sampling and learning approaches.}\label{tab:ode_estimation_comparison}
        \centering
        \begin{tabular}{|l|c|c|}
        \hline
        Method                                                                        & \begin{tabular}[c]{@{}l@{}}Total \#trajectories\end{tabular} & MSE   \\ \hline
        Ball sampling                                                                 & 150                          & 32.17 \\ \hline
        ROA sampling                                                                  & 73                           & 18.07 \\ \hline
        \begin{tabular}[c]{@{}l@{}}ROA sampling + \\ Lyapunov Regulrizer\end{tabular} & 73                           & 11.32 \\ \hline
        \end{tabular}
        \vspace{-2ex}
\end{table}

The outcome of the ODE identification using three different sampling and regularization approaches is shown in Figure~\ref{fig:vector_fields}. In each figure, it is shown a comparison between the flow of the true vector field (red) and the identified one (blue). The first column corresponds to sampling initial states without any knowledge of the Lyapunov function or ROA. The second column corresponds to the learned ODE when the estimated ROA is only used for cleverly sampling the initial states. In the ball sampling, the learned vector field is drastically different from the true one in the central regions of the plot that corresponds to the ROA. This shows the inefficiency of this sampling method when the accuracy of the learned vector field within the ROA is of concern. The second plot shows a clearly better match in the areas within the ROA compared with ball sampling. Notice that the mismatch between the learned and true vector fields outside the ROA (in the corners of the plots) are expected since almost no trajectory will explore those regions since initial states are always sampled around the ROA.

The third column corresponds to the scenario when the estimated ROA is used for clever sampling and the Lyapunov function is used for regularization as in~\eqref{eq:Lyapunov_regularizer}. The quantitative comparison of these three approaches is shown in Table~\ref{tab:ode_estimation_comparison} in terms of the mean squared error (MSE) between the estimated and the true vector field within the state space region consisting of the true ROA.
As the table shows, clever sampling using the ROA knowledge achieves superior results with fewer experiments. Besides, using the information of the Lyapunov function as a regularizer decisively improves the estimate of the ODE.

\section{Conclusion}
A method to co-learn the ROA and ODE of a dynamical system from the observed trajectories is proposed. Multilayer perceptrons are used to learn each component based on iterative supervised training. The algorithm approximates the true ROA as the maximal contractive level set of a Lyapunov function, while the ODE is learned by minimizing a regressor-type loss function. Crucially, the ROA is used to formulate the second type of loss function which regularizes the ODE fitting problem by endowing it with this local stability constraint.
Knowledge of the estimated ROA also enables a better sample complexity of the algorithm by informing the selection of the trajectories used for the purpose of training.

\bibliographystyle{IEEEtran}

\bibliography{main}             

\end{document}